\documentclass[letterpaper, 10 pt, conference]{ieeeconf}
\IEEEoverridecommandlockouts

\usepackage{amsmath,amsfonts}
\usepackage{algpseudocode}
\usepackage{algorithm}
\usepackage{array}
\usepackage[caption=false]{subfig}
\usepackage{tabularx}
\usepackage{colortbl}
\usepackage{textcomp}
\usepackage{stfloats}
\usepackage{multirow}
\usepackage{verbatim}
\usepackage{graphicx}
\usepackage{cite}
\usepackage{flushend}
\usepackage{diagbox}
\usepackage[dvipsnames]{xcolor}
\usepackage{threeparttable}
\usepackage{booktabs}
\usepackage{empheq}
\usepackage{makecell}
\usepackage{url}
\usepackage{soul}
\usepackage{tikz}
\let\labelindent\relax
\usepackage{enumitem}
\usepackage{gensymb}
\usepackage[colorlinks=true, linkcolor=black, citecolor=black, urlcolor=blue]{hyperref}
\definecolor{redvw}{HTML}{F36B9C}
\definecolor{bluevw}{HTML}{1E88E5}
\definecolor{greenvw}{HTML}{005648}

\begin{document}
\title{See, Plan, Cut: MPC-Based Autonomous Volumetric Robotic Laser Surgery with OCT Guidance}
\author{Ravi Prakash \textsuperscript{1*$^{\dagger}$},
Vincent Y. Wang \textsuperscript{1*},
Arpit Mishra \textsuperscript{1},
Devi Yuliarti \textsuperscript{1},
Pei Zhong \textsuperscript{1},\\
Ryan P. McNabb \textsuperscript{2},
Patrick J. Codd\textsuperscript{1,2$\ddagger$},
Leila J. Bridgeman\textsuperscript{1$\ddagger$} % No comma here
% --- Use negative space here to pull the link UP ---
% --- Footnotes ---
\thanks{$^{*}$Equal contribution. $^{\ddagger}$Equal advising. }%
\thanks{\href{https://raprakashvi.github.io/see-plan-cut/}{Project Website}}
\thanks{Code: \href{https://github.com/raprakashvi/see_plan_cut}{\texttt{github.com/raprakashvi/see\_plan\_cut}}}\thanks{$^{1}$Duke University, Durham, NC, USA.}%
\thanks{$^{2}$Duke University School of Medicine, Durham, NC, USA.}%
\thanks{$^{\dagger}$Corresponding author: {\tt\small ravi.prakash@duke.edu}}
}

\maketitle

\begin{abstract}
Robotic laser systems enable sub-millimeter, non-contact tissue resection, yet existing platforms lack volumetric planning and intraoperative feedback. We present \textbf{RATS (Robot-Assisted Tissue Surgery)}, an intelligent optical coherence tomography (OCT)-guided robotic platform for autonomous volumetric soft tissue resection. RATS integrates macro-scale RGB-D imaging, micro-scale OCT, and a fiber-coupled surgical laser, calibrated through a novel multistage alignment pipeline that achieves OCT-to-laser calibration accuracy of $0.161 \pm 0.031$~mm. A super-Gaussian laser--tissue interaction (LTI) model characterizes ablation morphology with an average RMSE of $0.231 \pm 0.121$~mm, outperforming Gaussian baselines. A sampling-based model predictive control (MPC) framework operates directly on OCT voxel data to generate closed-loop, constraint-aware resection trajectories, achieving $0.842$~mm RMSE (root-mean-square error) and improving intersection-over-union agreement by $64.8\%$  compared to feedforward execution. RATS also detects and preserves subsurface structures, demonstrating the first closed-loop autonomous volumetric robotic laser resection with OCT guidance. To our knowledge, this is the first demonstration of closed-loop autonomous volumetric robotic laser resection with OCT guidance, enabling precise, obstacle-aware tissue removal with potential in neurosurgery.
%applications in neurosurgery and complex soft tissue oncology.

\end{abstract}
%%%###################################################################################
% \begin{IEEEkeywords}
% Robotic Surgery, Volumetric Resection, Laser–Tissue Interaction (LTI), Optical Coherence Tomography (OCT), Model Predictive Control (MPC)
% \end{IEEEkeywords}
%%%###################################################################################
\section{Introduction}
%%%###################################################################################
Robot-assisted surgery has significantly improved precision, dexterity, and safety in minimally invasive procedures~\cite{moustris2011evolution,fiorini2022concepts}, largely through the integration of novel instruments with complementary sensing and control. However, most systems still rely on teleoperation for critical tasks like tumor resection, which requires accurate volumetric removal while preserving critical structures—particularly in neurosurgery with irregular margins and dense neurovasculature. 

Laser-based scalpels offer sub-millimeter resolution~\cite{lee2022end}, non-contact energy delivery, and simultaneous ablation and coagulation~\cite{vogel2003mechanisms,lee2022laser}, making them ideal for precise soft tissue removal. However, their inherent precision makes them challenging to dexterously wield during surgeries, making adoption difficult. Robotic systems have demonstrated progress in laser manipulation~\cite{york2021microrobotic}, sensing~\cite{gunalan2023towards,ma2025tumormap}, and control~\cite{li2024auto} for ablation tasks, but remain constrained by three major limitations: (i) lack of integrated volumetric sensing, (ii) poor image-to-laser calibration, and (iii) absence of model-based constraint-aware planning. As a result, current approaches rely on open-loop raster scanning~\cite{ross2018optimized} or simulation-only predictive planning~\cite{wang2024sampling}, limiting clinical utility in complex 3D surgical environments.  

\begin{figure}[] % requires \usepackage{float}
    \centering
    \vspace{-2mm} % reduce space above
    \includegraphics[width=0.8\linewidth]{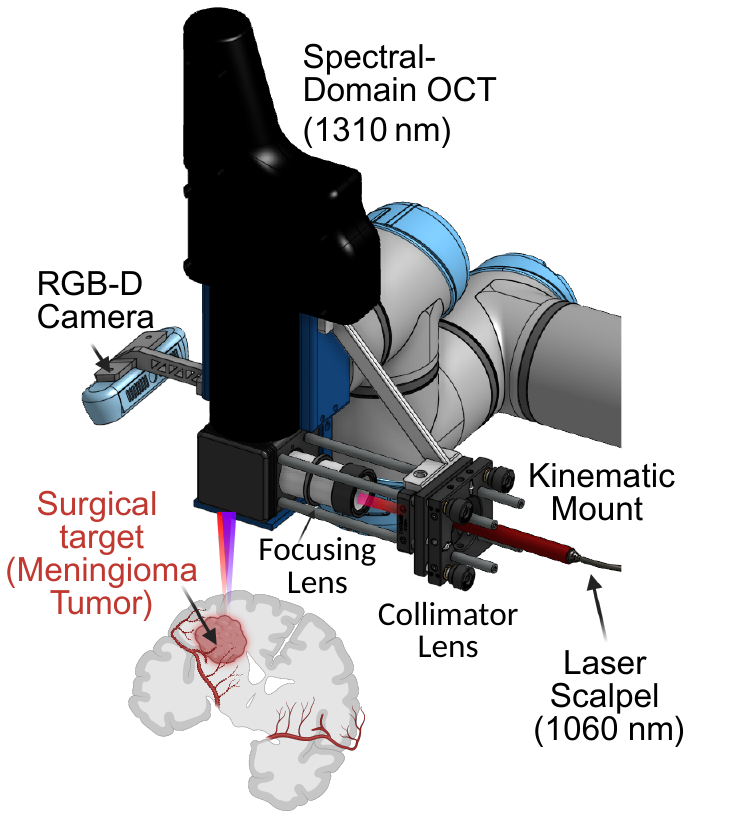}
    \vspace{-2mm} % reduce space below
    \caption{\textbf{RATS Platform:} Hardware configuration for OCT-guided volumetric laser surgery: The robot-mounted platform integrates a spectral-domain OCT scanner (1310 nm) and fiber-coupled surgical laser (1060 nm), co-axially aligned with OCT through a dichroic mirror. A collimated laser scalpel beam is focused through an achromatic doublet-based focusing lens. An RGB-D camera provides macro-scale information over OCT's micro-scale features. The system is shown targeting a simulated meningioma tumor embedded within vascular tissue, representative of neurosurgical environments requiring sub-millimeter precision and constraint-aware planning \cite{biorender_fig1}}.
    \label{fig:system_overview}
    \vspace{-15pt}
\end{figure}
%%%%############################################################################
Recent work has leveraged high-resolution imaging modalities for crater reconstruction and subsurface feature detection~\cite{ma20233d,Lan:25}, such as optical coherence tomography (OCT)~\cite{huang1991optical}, which provides micron-scale resolution several millimeters deep. When paired with a 6-DoF robotic arm and a laser resection tool, OCT enables intraoperative volumetric planning~\cite{prakash2025portable,ma2025robotics}. However, closed-loop robotic laser resection requires coordination between an accurate LTI model, a constraint-aware volumetric planner, and precise calibration between imaging and actuation to achieve accurate surgical resection. Such closed-loop integration is especially critical in neurosurgical oncology, such as in meningioma resection, where tumors may encapsulate critical vasculature and manual resections are both time-consuming and high-risk.  
With these challenges in mind, we introduce RATS (Robot-Assisted Tissue Surgery), a novel OCT-guided opto-mechanical robotic platform for intelligent and precise volumetric soft-tissue resection (Fig.~\ref{fig:system_overview}). RATS integrates OCT-guided 3D imaging with a calibrated fiber-coupled surgical laser and a sampling-based model predictive control (MPC) planner for constraint-aware resection. Our contributions are:

\begin{itemize}[left=0pt]
\item \textbf{Novel Surgical Platform:} A modular RATS system integrating imaging (OCT and RGB-D) with an optically coupled co-axial fiber-coupled laser scalpel, calibrated via a multi-stage pipeline to align OCT, laser, and end-effector frames with sub-millimeter accuracy.
\item \textbf{LTI modeling:} A data-driven, OCT-guided LTI model validated on tissue phantoms, using a super-Gaussian laser beam formulation to capture ablation crater morphology and tissue response.  
\item \textbf{Volumetric Resection Planning:} A sampling-based MPC framework operating directly on OCT voxel data, capable of generating constraint-aware resection trajectories and executing closed-loop ablations.
\end{itemize}

Prior work has addressed isolated components of this pipeline but remains limited to simulation, single-point, or line-based resection. Ross et al.~\cite{ross2018optimized} proposed a genetic algorithm for raster-based surface ablation, though their method was limited to 2D surface evaluation and lacked sub-surface awareness. Wang et al.~\cite{wang2024sampling} proposed MPC-based planning; their system was validated only in simulation and did not integrate real-time sensing. Acemoglu et al.~\cite{acemoglu2017laser,ma20233d} introduced data-driven LTI models for single-point or line-based ablations but lacked full volumetric integration. Prakash et al.~\cite{prakash2025portable} demonstrated a dual-sensor robotic platform for large-area OCT 3D volume reconstruction, though without integrated planning or resection.

To our knowledge, this is the \textbf{first work to demonstrate OCT-guided autonomous volumetric robotic laser resection} in anatomically relevant settings with closed-loop feedback. RATS tightly integrates perception, planning, and actuation for large-area (centimeter-scale) subsurface-aware (millimeter-scale) laser surgery, advancing the frontier of surgical robotics with potential applications in neuro-oncology, head and neck tumor removal, and soft tissue sarcoma resection requiring precision and obstacle avoidance.
%%%###################################################################################
%In this work, we present RATS (Robot-Assisted Tissue Sugery) (RATS), a novel opto-mechanical robotic surgical system designed for intelligent and precise volumetric soft tissue resection for advanced oncological care. RATS integrates OCT-guided volumetric imaging with an optically aligned and calibrated fiber-coupled surgical laser scalpel, utilizing an intelligent surface-constraint aware model predictive control (MPC)-based planning framework to execute safe and autonomous tissue resections over a large workspace. 
%tissue property estimation. 
%Why laser surgery -> Current gaps -> Need for OCT -> Lack of planning algorithm for volume removal
% To our knowledge, this is the first work to perform volumetric tissue resection using a robotic laser scalpel.
%%%###################################################################################

%%%################################################################################
\begin{figure}[!t]
    \centering
     {\includegraphics[width=0.90\linewidth]{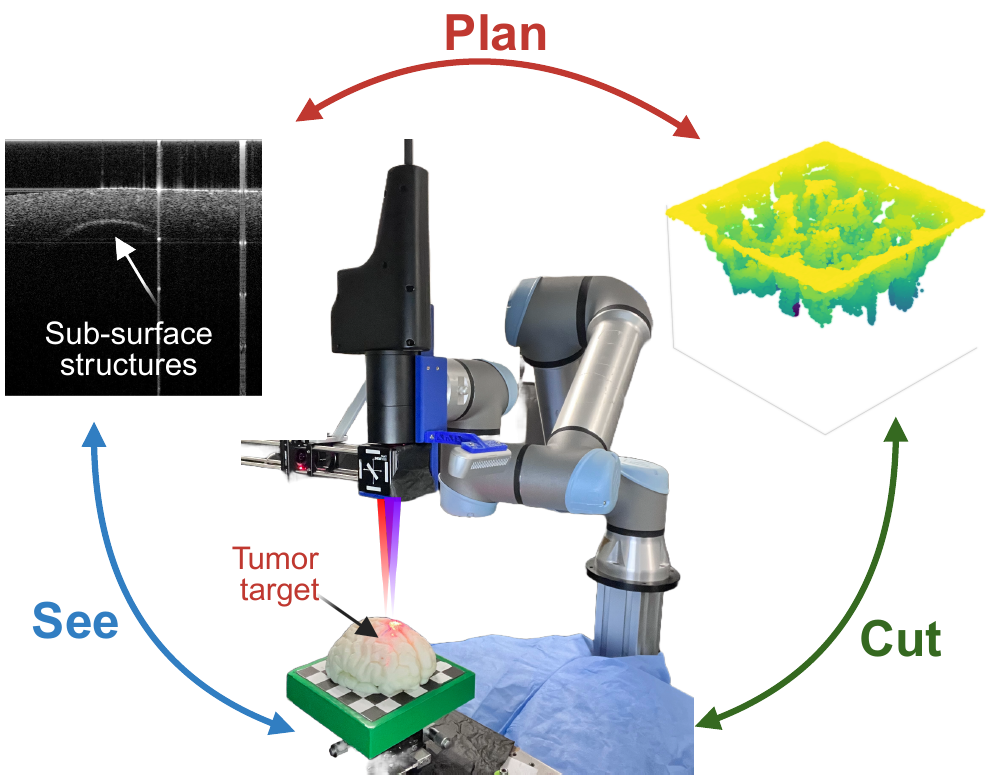}}
    \caption{\textbf{System Overview:} Experimental setup with a brain phantom for demonstration. The OCT sensor \textit{sees }the surface and sub-surface structures, and the reconstructed 3D structure is then used to generate a volumetric resection \textit{plan}. The planner generates a sequence of robot states and laser parameters to \textit{cut} the tissue with clinically relevant objectives.\cite{biorender_fig2}}
    \label{fig:rats_platform}
    \vspace{-10pt}
\end{figure}
%%%################################################################################

%%%###################################################################################
\section{Methods} \label{sec:methods}
%%%###################################################################################
\subsection{System Overview}\label{sec:sys_overview}
The RATS platform is a modular opto-mechanical robotic system designed for high-resolution imaging and precise, non-contact laser-based tissue resection (Fig.~\ref{fig:rats_platform}). It integrates:

\begin{itemize}[left=0pt]
    \item \textbf{Imaging stack:} A low-cost spectral-domain (SD) OCT system, 1310 nm, (Lumedica Inc, NC, USA), with a working distance of 110 mm,  is co-axially aligned with a near-infrared fiber-coupled surgical laser-scalpel using a dichroic mirror for synchronized imaging and ablation. A secondary RGB-D camera (Intel RealSense D435i) is used for coarse scene reconstruction and ROI selection. All components are mounted to a 6-DoF collaborative robotic arm with a precision of $\pm0.03$ mm (UR5e, Universal Robots, Denmark), which provides repeatable pose control for multi-angle scanning, laser delivery, and end-effector motion during volumetric resections.
    
    \item \textbf{Laser optics and scalpel assembly:} A surgical laser ($\lambda$ =1060 nm, 10 W, multi-mode continuous wave, Crystal Laser, NV, USA) is collimated (5 $\pm$ 1 mm) and focused using an achromatic doublet focusing lens (f = 75\,mm, AC254-075-C Ø1", Thorlabs, NJ, USA). A long-pass dichroic mirror with an edge wavelength of 1072.4 nm (LPD02-1064RU-25x36x2.0, IDEX Semrock, IL, USA) co-aligns the OCT and laser scalpel paths, and a $\lambda/4$ UV fused silica Anti-Reflection (AR) coated protective window (TECHSPEC, Edmund Optics, NJ, USA) is installed at the exit port of the dichroic cage to protect the optics from debris generated during ablation. Paper burn tests are employed to measure the spot size at the focal length to be 0.9 mm. Total energy delivered over a set period of time under various duty cycles is measured using a power meter (Table~\ref{tab:duty_vs_energy}).

   \item \textbf{Phantom development:} The laser scalpel and OCT have different wavelengths, so a tissue-mimicking phantom was developed using 2\% (w/w) agarose for matrix, 10\% (w/w) intralipid for OCT optical scattering, and 0.1\% (w/w) India-Ink as a chromophore to absorb 1060\, nm laser energy. Unlike other brain tissue mimicking phantoms focusing on mechanical properties \cite{prakash2023brain}, this phantom simulates the optical properties of soft tissue and is OCT-visible, with the addition of India Ink to promote absorption at the laser wavelength \cite{tucker2018creation}.

    \item \textbf{Controls architecture:} A Python-based control architecture manages robotic positioning, OCT scanning, laser trigger control, and closed-loop feedback. The interface supports pre-programmed patterns, trajectories, and voxel-level data from OCT, used to compute tissue resection paths in simulation and execution. 
\end{itemize}

\begin{table}[t]
\centering
\caption{Mean and standard deviation of measured laser energy.}
\label{tab:duty_vs_energy}
\resizebox{\columnwidth}{!}{%
\begin{tabular}{@{}rccccccccc@{}}
\toprule
\textbf{Duty Cycle (\%)} & 20 & 30 & 40 & 50 & 60 & 70 & 80 & 90 & 100 \\
\midrule
\textbf{Mean Energy (J)} & 1.91 & 3.50 & 5.11 & 6.75 & 7.68 & 8.29 & 9.31 & 10.03 & 11.05 \\
\textbf{Std. Dev. ($\times 10^{-2}$ J)} & 1.6 & 1.4 & 3.3 & 3.0 & 6.4 & 3.4 & 12.0 & 10.3 & 23.7 \\
\bottomrule
\end{tabular}%
}
\vspace{-10pt}
\end{table}

\subsection{Multi-Sensor Calibration}
\begin{figure}[!t]
    \centering
    \includegraphics[width=0.95\linewidth]{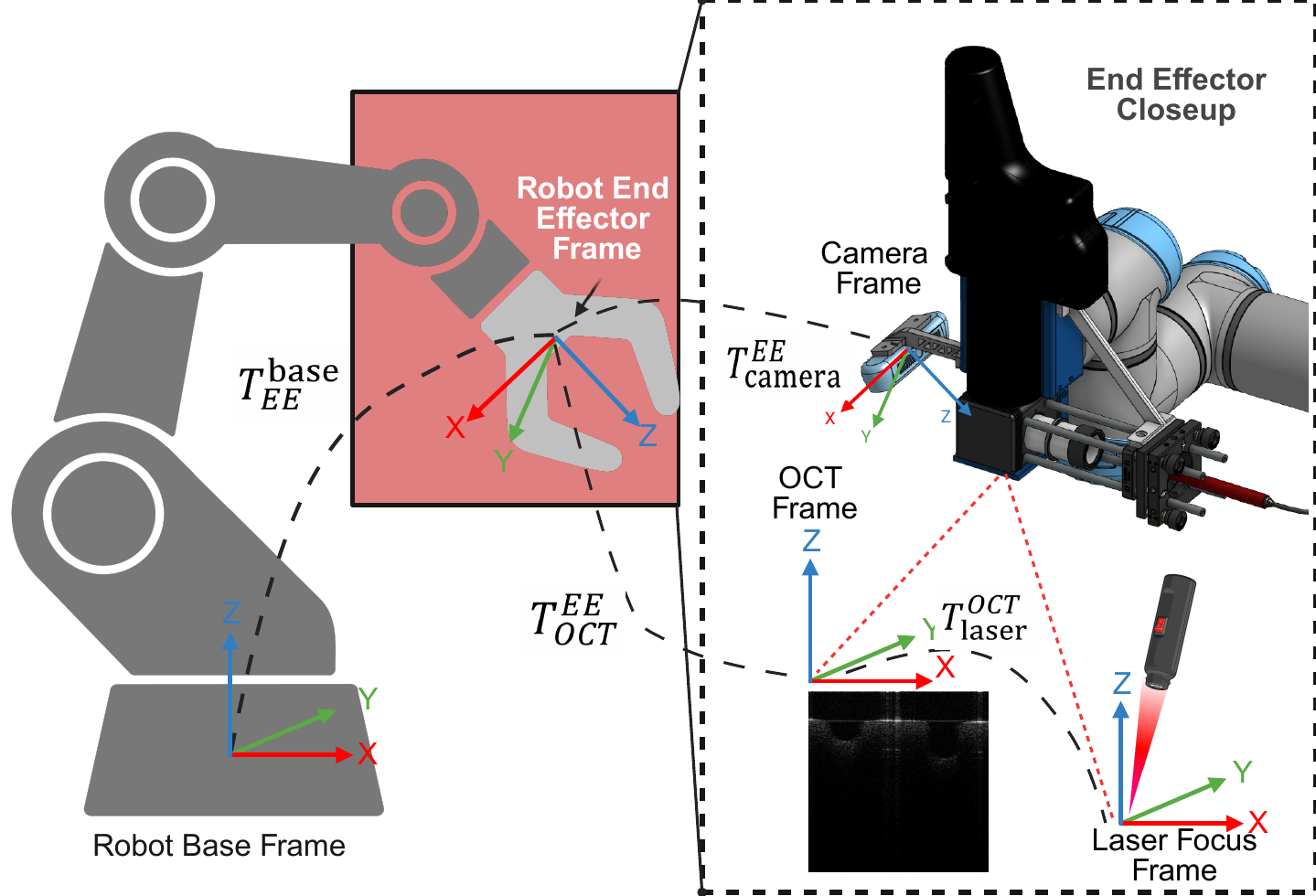}
    \caption{\textbf{System calibration pipeline and coordinate frame hierarchy:} The robot end-effector houses the OCT, RGB-D camera, and laser scalpel. Transformations between robot base, end-effector, OCT, camera, and laser focus frames are shown. Includes OCT 2D-to-3D volume registration, OCT-to-end-effector (EE) frame transformation, and laser-to-OCT alignment. Dashed lines indicate calibrated extrinsic transforms (\( T^{EE}_{\text{OCT}} \), \( T^{\text{OCT}}_{\text{laser}} \), \( T^{EE}_{\text{camera}} \)). The figure illustrates the frame alignments that are critical for precise imaging and laser actuation.}
    \label{fig:calibration}
    \vspace{-15pt}
\end{figure}

Accurate volumetric resection requires precise alignment between the robot, imaging (OCT) and laser-ablation modules. We implement a multi-stage calibration procedure to estimate the rigid transformations necessary for closed-loop control and voxel-level accurate targeting (Fig.~\ref{fig:calibration}). 
%includes several components, such as RGBD camera calibration, system calibration of OCT, OCT to robot end-effector (EE), and laser scalpel to OCT calibration. 

\subsubsection{Camera to End-effector Calibration}
We estimate the extrinsic transformation between the RGB-D camera and the robot end-effector, $T^{EE}_{\mathrm{camera}}$, using the OpenCV hand–eye calibration toolbox. A standard checkerboard is mounted in the workspace and observed from multiple poses. With the camera intrinsics pre-calibrated, the $AX = XB$ formulation~\cite{tsai2003versatile} is used to solve for $X = T^{EE}_{\mathrm{camera}}$, where $A$ corresponds to robot motion and $B$ to camera-observed checkerboard motion. 
%Camera extrinsic calibration is a well-studied problem, and multiple libraries exist to perform the task. An RGBD camera is used to guide the robot to the target when needed. 

\subsubsection{OCT to End-Effector Calibration}
Before 3D OCT volumes (C-scans) can be interpreted in world coordinates, the pixel-to-millimeter scale must be determined. 
%spacing values for converting OCT images to coordinate values is needed. 
To estimate lateral and axial scaling factors, we scan fiducial markers of known dimensions placed at the OCT focal plane.
%Fiducial markers of known lengths were placed at the focal length of the OCT, and the distance between two markers was measured for lateral and axial axis following the method described in \cite{}.
To register the OCT frame to the robot end-effector, we estimate the transformation $T^{EE}_{OCT}$ using the $AX = XB$ formulation~\cite{tsai2003versatile,ma2025geometry}. For any robot orientation, $i$, the matrix, $^{(i)}T^{\mathrm{world}}_{EE}T^{EE}_{OCT} {}^{(i)}T^{OCT}_{\mathrm{marker}}$, is constant as it relates a point in a stationary frame relative to the calibration pattern to its coordinates in the world frame. Hence, obtaining multiple values of $T^{\mathrm{world}}_{EE}$ and $T^{OCT}_{\mathrm{marker}}$ and rearranging them into $A$ and $B$ matrices provides an estimate of $X=T^{EE}_{OCT}$ \cite{tsai2003versatile}.

While the $T^{\mathrm{world}}_{EE}$ matrix is easily obtained from forward kinematics, our method for estimating the $T^{OCT}_{\mathrm{marker}}$ matrix uses a two-dot calibration board paralleling \cite{ma2025geometry}.
%(Fig.~\ref{fig:calibration_error}).
We define the stationary ``marker'' frame by denoting the vector between the two dots as the $+X$ axis, while the vector orthogonal to the marker plane is denoted as the $+Z$ axis. The $+Y$ axis can then be obtained from a cross product between the two vectors, which allows us to construct the OCT to marker transformation ($T^{OCT}_{\mathrm{marker}}$).

\subsubsection{Laser to OCT Extrinsic Calibration}
\begin{figure}[h!]
    \centering
     \includegraphics[width=0.65\columnwidth]{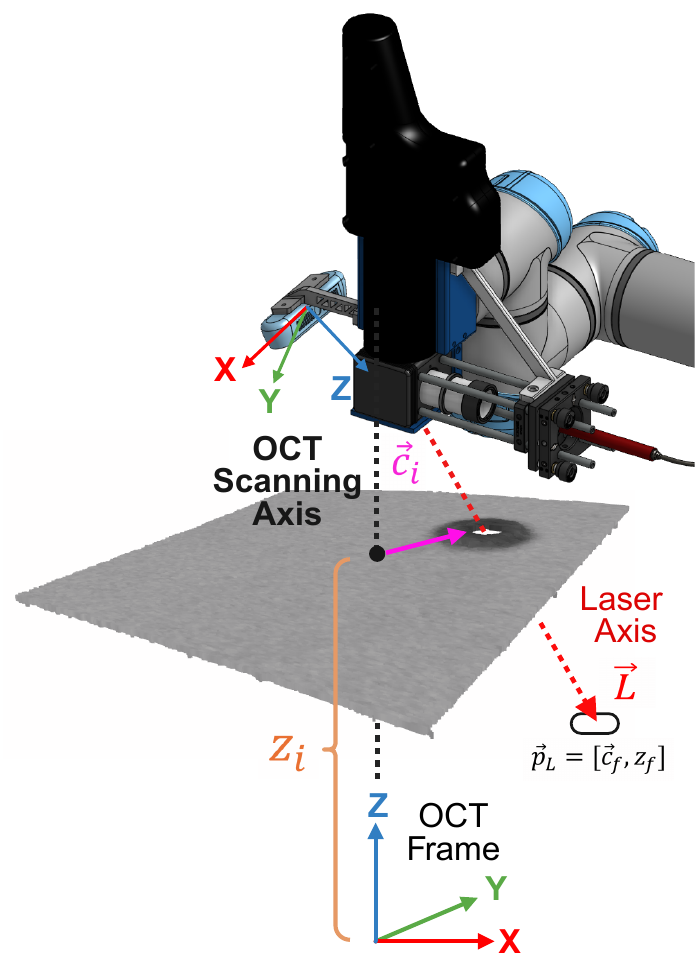}
    \caption{\textbf{Laser Calibration Method:} Calibration schematic showing OCT frame (blue--green--red), OCT scanning axis, and laser axis alignment. The intersection point defines the laser focal point $\vec{p}_{L}$, estimated from crater centers $\vec{c}_{i}$ at different depths $z_{i}$.}
    \label{fig:laser_calib}
    \vspace{-10pt}
\end{figure}

To track the laser, we calibrate directly to the OCT system by estimating $T^{OCT}_{\mathrm{laser}}$. This matrix can be directly estimated by obtaining the vector $\vec{L}$ that points along the laser axis in the OCT frame, and the location of the laser focal point, $\vec{p}_L$. This can be obtained by performing a sequence of ablations at different heights, $z_i$, along the OCT axis, then estimating the center $\vec{c}_i=[x_i, y_i]$ of each ablation crater on the $XY$ plane in the OCT frame. These points form a parametric line in $\mathbb{R}^3$ representing the laser axis, $\vec{L}$, which can be obtained from calculating a line-of-best-fit through collected $\vec{c}_i$ and $z_i$ pairs. To estimate $\vec{p}_L$, we can find the point along the line $\vec{L}$ corresponding to a known focal distance, $z_f$ (Fig.~\ref{fig:laser_calib}).

\subsection{Laser Tissue Interaction Model} \label{sec:LTI_model}
%%%######################################################
%%%%%%%##################################################
To generate a tissue ablation plan, the tissue's response to varied geometry and energy distribution of laser-induced ablation craters is needed. We model single-point tissue resection using a two-dimensional Gaussian beam model \cite{ross2018optimized,wang2024sampling} and a steady-state ablation formulation \cite{vogel2003mechanisms}. This enables us to estimate interpretable tissue parameters, such as tissue density and enthalpy, allowing us to predict ablation craters.

\paragraph*{Gaussian profile estimation and interpretation}

The 3D ablation crater can be represented geometrically by conditioning the steady state laser ablation model upon the laser beam position. With the incident laser beam centered at ($\mu_x, \mu_y)$ on the tissue surface, the resulting ablation crater depth, $f$, at point, $(x,y)$, on the surface can be calculated as

\begin{figure*}[t]
    \centering
    \includegraphics[width=0.9\linewidth]{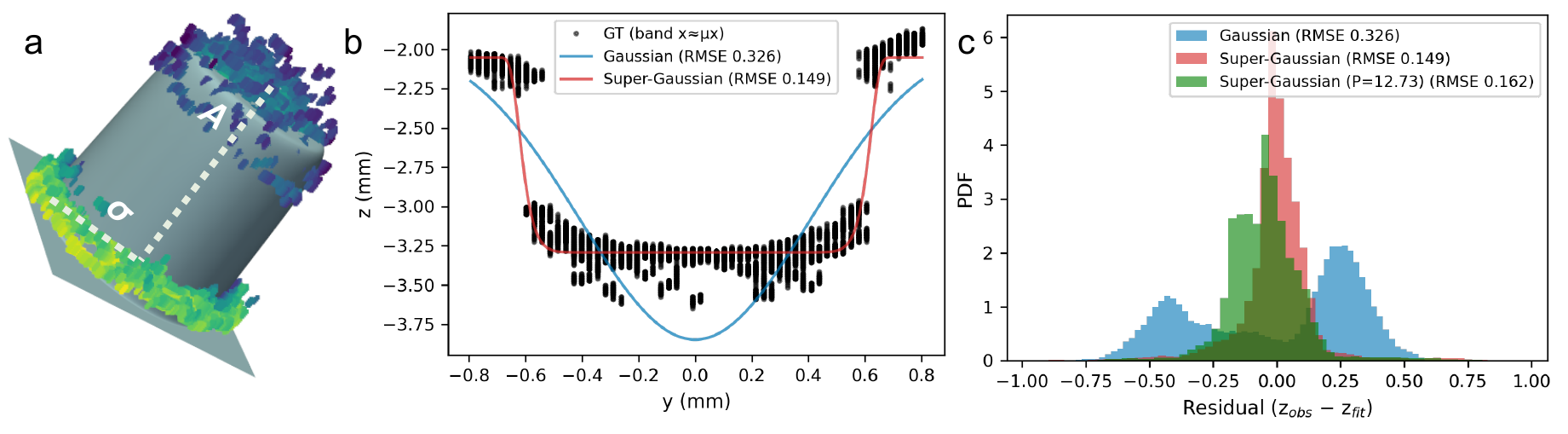}
    \caption{\textbf{Laser--tissue interaction (LTI) model estimation.} 
    \textbf{(a)} Example of super-Gaussian parameter estimation through fitting a curve to the ground-truth point cloud obtained from OCT B-scans. This process enables estimation of single-point tissue response for a given laser energy. 
    \textbf{(b)} Example of Gaussian and super-Gaussian curve fits on a ground-truth point cloud along the $x$-axis. Gaussian curve estimation does not capture the multimode beam response adequately. 
    \textbf{(c)} Probability density function of residual error from curve fitting ($z_\text{fit}$) with respect to observed ($z_\text{obs}$) point cloud for Gaussian and super-Gaussian fits. The red residual plot represents error from a super-Gaussian fit where $P$ was allowed to vary as a parameter to be fitted, while the green residual plot represents error from a super-Gaussian fit where an average $P$ across all fitting experiments was passed in as a constant parameter ($P = 12.73$).}
    \label{fig:lti_model}
    \vspace{-15pt}
\end{figure*}

\vspace{-16pt}
\begin{align}
\label{eq:supergauss}
f(x,y) =&  - {A}\mathrm{max}\left(0, E
\exp\left[-\frac{1}{2}\left(\dfrac{r^{2}}{\sigma^{2}}\right)^{P}\right]-\phi\right), \\
r^2=&(x-\mu_x)^2+(y-\mu_y)^2,% [8pt]
% P = 1 \quad \text{(Gaussian beam)} \\ 
% P > 1 \quad \text{(Super-Gaussian beam)}\\
% P \rightarrow \infty \quad \text{(flat disk of radius $\sigma$)}
% \end{array}
\end{align}
where $A$ is an amplitude dependent on laser/tissue parameters, $\sigma$ is the standard deviation of the distribution, $\phi$ is an offset parameter representing the minimum energy required for cutting to occur, $E$ is the total energy delivered, and $P\in \mathbb{R}$ modulates the profile sharpness. When \( P=1 \), the profile reduces to a standard Gaussian, which was used in \cite{ross2018optimized, wang2024sampling, pacheco2024automatic}. For \( P > 1 \), the crater has a flatter bottom and steeper sidewalls, which is often observed in high-power, multimodal fiber with quasi-top-hat beams (Fig.~\ref{fig:lti_model}) such as ours.

\subsection{Parameter regression}
% To perform the model fitting to the super-Gaussian curve,
For fitting the model to the super-Gaussian curve, the LTI parameters $A$, $\sigma$, and $P$ in Equation~\ref{eq:supergauss} are estimated via the Levenberg–Marquardt algorithm \cite{gavin2019levenberg}, adaptively interpolating between gradient descent and Gauss–Newton updates for efficient convergence in nonlinear parameter spaces. The influence of edge noise and overfitting is reduced by fitting to a circular region within \( 2\sigma \) of the crater center. Given cutting errors, $e_i= z_i - f(x_i, y_i)$, the model's accuracy is evaluated using the RMSE,
% root mean square error (RMSE), 
$N^{-0.5} (\sum_{i=1}^{N} e_i^2 )^{0.5}$, overcut (\%OC), $\sum_{i=1}^{N} \mathrm{min}\left(0, e_i\right) / \sum_{i=1}^{N} e_{i,0}$, undercut (\%UC), $\sum_{i=1}^{N} \mathrm{max}\left(0, e_i\right) / \sum_{i=1}^{N} e_{i,0}$, and intersection over union (IoU), 
\begin{align*}
    \text{IoU}=\frac{V_{real} \cap V_{fit}}{V_{real} \cup V_{fit}}
    =\frac{ \sum_{i=1}^{N} \mathrm{max}( z_i , f(x_i,y_i) ) }{ \sum_{i=1}^{N} \mathrm{min}( z_i , f(x_i,y_i) )  },
\end{align*}
where $z_i$, $f(x_i,y_i)$, and $z_{i,0}$ denote the ablation depth, target depth, and initial depth at $x_i, y_i$,  and $e_{i,0}=z_{i,0}-f(x_i,y_i)$.
Special importance is given to \% overcut, as ablated tissue cannot be ``uncut'', meaning overcut leads to difficulty and unremovable error in future planning. 

\subsection{Volumetric Resection Path Planning Algorithm} \label{sec:planner}
%We present the first experimental implementation of a closed-loop, constraint-aware volumetric resection planner capable of executing volumetric resection in both feedforward and feedback modes. The implementation presented in this work accounts for several practical challenges of non-contact, laser-based surgery, such as the inevitable presence of debris, geometric workspace constraints, imaging system artifacts, and estimation of performance error. 
%The primary imaging sensor, OCT, is an inherently discrete imaging modality, so 
%%\subsubsection{Sensing and discretization}
%OCT is used as the primary feedback sensor due to its micron-scale resolution and robustness in soft tissue imaging. OCT volume workspace is discretized, including the target (tumor) and hard constraints (critical structures to avoid). Clough-Toucher splines are used to interpolate between points and generate smooth surfaces.

%%\subsubsection{Ablation model and crater simulation.}
The planner (Alg.~\ref{alg1}) employs the sampling algorithm from \cite{wang2024sampling} and a single-shot laser–tissue interaction model to select ablation inputs $\vec{u} = (\vec{x}_L, \vec{\theta}_L, D)$, where $\vec{x}_L = (\mu_x, \mu_y)$ defines the laser incident point on the tissue surface, $\vec{\theta}_L = (\theta_x, \theta_y)$ specifies the laser incident angle, and $D$ is the laser duty cycle. 
% The planner employs the sampling algorithm from \cite{wang2024sampling} to select ablation inputs based on a single-shot laser–tissue interaction model (Alg.~\ref{alg1}). 
\begin{algorithm}
\caption{Sampling-Based MPC for Laser Resection}
\label{alg1}
\begin{algorithmic}
\State Inputs $\gets$ \texttt{surface, objective, constraint}
\\\hrulefill
\State cost $\gets$ $C(\texttt{surface, objective})$
\State Graph $\gets$ Add [\texttt{surface, cost}] as root node to tree
\Repeat
    \State currentNode $\gets$ randomly sample a node from \texttt{Graph}
    \State laser $\gets$ randomly sample inputs, ($\left[X_L, \Theta_L, P_L\right]$)
    \State nextSurf $\gets$ simulate ablation with (\texttt{currentNode}, \texttt{laser})
    \State nextCost $\gets$ $C(\texttt{nextSurf, objective})$
    \If {\texttt{nextNode} does not violate \texttt{constraint}}
        \State Graph $\gets$ add \texttt{nextNode}
    \EndIf
\Until{number of nodes in \texttt{Graph} exceeds $k_F$}
\State \Return node with the lowest cost within \texttt{Graph}, and path from \texttt{root} to lowest cost node containing all inputs
\end{algorithmic}
\end{algorithm}
The algorithm builds a tree whose edges contain inputs and nodes contain objective surfaces (tumor to remove), constraint surfaces (critical regions to avoid), and associated costs
\begin{align}
    C(\mathcal{X}) &= \sqrt{\frac{1}{N}\sum_{i=1}^N\left(\mathrm{min}\left(0, e_i\right)+\lambda\mathrm{max}\left(0, e_i\right)\right)^2},%\\
    %D_i &= z_i - f_o(x_i, y_i)
\end{align}
% with weighting factor $\lambda\geq1$ to favor undercuts, and edges containing candidate inputs. The inputs are parameterized as $\vec{u} = (\vec{x}_L, \vec{\theta}_L, D)$, where $\vec{x}_L = (\mu_x, \mu_y)$ defines the laser incident point on the tissue surface, $\vec{\theta}_L = (\theta_x, \theta_y)$ specifies the laser incident angle, and $D$ is the laser duty cycle.
with weighting factor $\lambda\geq1$ to favor undercuts, and edges containing candidate inputs. The root node is obtained either from OCT scanning or preoperative imaging. Child nodes/edges are obtained by sampling possible parent nodes and predicting the resultant surface after applying a randomly sampled input via \autoref{eq:supergauss}. 
%
%%##################################################################
% . To build the tree, a node is sampled as the initial node by randomly selecting an existing node in the tree with low objective cost, where the objective cost of the state in node $i$ is defined as
% % \begin{align}
% %     C_i(\mathcal{X}) &= \sqrt{\frac{1}{n}\sum_{i=1}^n\left(\mathrm{min}\left(0, D_i\right)+\lambda\mathrm{max}\left(0, D_i\right)\right)^2}\\
% %     D_i &= z_i - f_o(x_i, y_i),
% % \end{align}
% a modified RMSE cost function. 
%The parameter $\lambda\geq1$ assigns higher penalty to overcut regions where the tumor has been exceeded and healthy tissue has been removed, as there is no way of `replacing' healthy tissue and thus its removal should be preferentially avoided. 
%%##################################################################
%Each input is selected at random from the set of allowable laser positions, angles, and duty cycles. The resulting child node contains the simulated tissue surface generated by applying the selected laser cut to its parent node's surface. 
If the resulting surface violates any constraints, the node is not added, guaranteeing constraint adherence. This procedure repeats for a set number of iterations $(k_F)$, before an input sequence is chosen as the path between the root node and the lowest-cost node in the tree. %This process is summarized in Figure~\ref{fig: graph-diagram} and Algorithm~\ref{alg1}. 
In feedforward mode, this algorithm is repeated, with the prior end-node as the root node of a new tree, whereas in feedback mode a new OCT scan provides an observed surface for the new root node.
%%##################################################################%%##################################################################
%until a target amount of tumor is removed. Additionally, the algorithm is capable of incorporating feedback in a sense-act loop by performing a sequnce of $m$ cuts, before rescanning with OCT to obtain the true surface boundary and replanning for another $m$ cuts. 

% \textbf{Something is missing: link between objective cost function and the model. How is the target defined? Some inputs on the planning horizon, how convergence was addressed in the simulations, and how the feedback mode operates. The distinction between feedback and no-feedback cases also needs to be clarified}

% A sequence of inputs is chosen to lead to the lowest objective cost, defined as the RMSE between the tissue boundary and the desired boundary. This sampling loop continues until the desired volume of tumor is resected, or no feasible inputs remain within the safety constraints, such as critical structures or simulated blood vessels or anatomical structures.

%\begin{enumerate}
%    \item Cite \cite{wang2024sampling} and mention the algorithm being used here. 
%    \item modifications made for python implementation, state variables, role of the parameter corresponding to an MPC prediction horizon and a discussion of the impact of its tuning in terms of control performance, 3) Cost function for the MPC algorithm must be clearly stated.
%\end{enumerate}
%%%######################################################

%%%%%%%##################################################
\section{Experiments and Results}
%%##################################################################
We evaluated the RATS platform, LTI model, and MPC-based planner to assess accuracy, robustness, and surgical applicability. Experiments included (i) system calibration, (ii) LTI model fitting, and (iii) volumetric resection on tissue phantoms and \textit{ex vivo} porcine with and without feedback.

%The evaluations spanned calibration accuracy, resection precision, and planning algorithm performance on tissue phantoms with and without feedback. \hl{Closed-loop feedback was applied to embedded subsurface critical structures and \textit{ex-vivo} tissue}.

\subsection{System Verification and Error analysis} \label{sec:error_analysis}

\subsubsection{OCT calibration } 
Using fiducial markers, the OCT pixel spacing was determined to be $14~\mu$m laterally and $14.59~\mu$m axially. For each C-scan, 256 B-scans were collected with a distance of $28~\mu$m between adjacent B-scans~\cite{ma2025geometry,prakash2025portable}. Each B-scan consisted of 512 A-scans, and the images were processed to obtain a 3D tissue surface emulating~\cite{ma20233d,prakash2025portable}. Each C-scan encompassed a $7.168 \times 7.168 \times 7.6288~\mathrm{mm}^3$ volume, with axial penetration limited to $2$--$4~\mathrm{mm}$ based on tissue optical properties. Fig.~\ref{fig:calibration_error}a summarizes the calibration error for the OCT-to-EE alignment. The mean reprojection errors for the OCT-to-EE calibration were $0.638$~mm for the origin and $0.625$~mm for the $x$-axis, with occasional outliers above $2$~mm (Fig.~\ref{fig:calibration_error}a). 
%illustrates the marker-based OCT frame definition, where the $X$-axis is aligned with fiducial markers and the $Z$-axis corresponds to the surface normal, enabling robust OCT-to-end-effector transformation.

\subsubsection{Laser Calibration Verification}
A $3 \times 3$ dot grid was scanned by the OCT to generate a rasterized en-face image, enhancing dark circular blobs. Blobs were segmented via thresholding and morphological operations, and their centroids were extracted and then mapped back to the metric OCT frame. The $z$-coordinate at each center was obtained from a global plane fit to the full point cloud. 

\begin{figure*}[!t]
    \centering
    \includegraphics[width=0.9\linewidth]{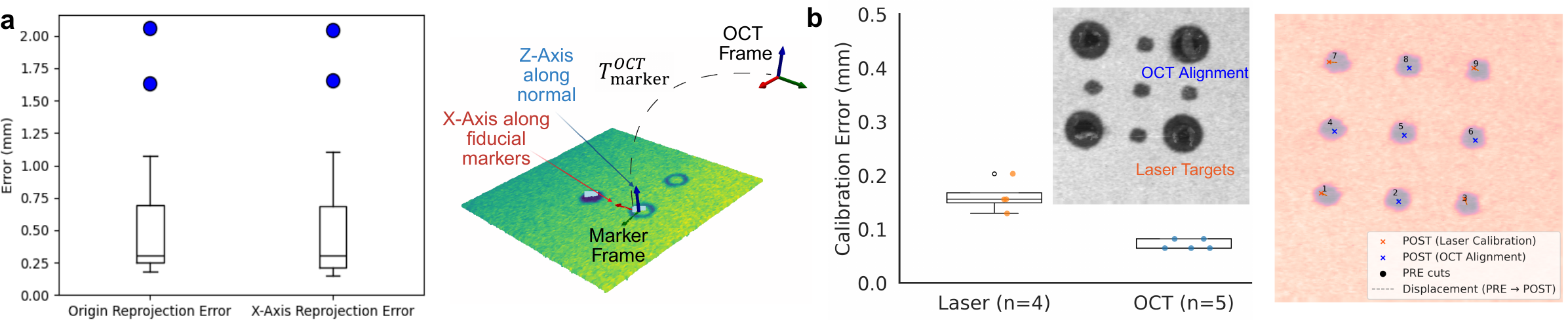}
    \caption{ \textbf{Calibration Error Analysis:} \textbf{(a)} OCT-to-end-effector calibration accuracy (left) and relevant coordinate frames (right). Note that the third, unlabeled fiducial marker is only used to differentiate between the positive and negative $Z$ direction of the otherwise symmetric OCT scan, and is not used in the calibration itself. \textbf{(b)} Laser-to-OCT calibration and representative \textit{en-face} image after laser calibration test (left), OCT-probe repositioning verification error (right).}
    \label{fig:calibration_error}
    \vspace{-10pt}
\end{figure*}

For laser scalpel calibration verification, four blobs were selected and ablated for $0.5$~s while maintaining orientation along the surface normal. The remaining targets were used to evaluate OCT repositioning error across successive scans (Fig.~\ref{fig:calibration_error}b). The measured laser calibration error was $0.161 \pm 0.031$~mm, well below the $0.5$~mm neurosurgical threshold, confirming calibration accuracy. Additionally, the OCT repositioning error to a previously scanned point was $0.071 \pm 0.009$~mm, further validating the calibration pipeline. The larger ablated circles observed in Fig.~\ref{fig:calibration_error}b arose from selectively higher absorption of the 1060~nm laser by black-dot chromophores, which enhanced local heat dissipation. 

\subsection{LTI model performance}
Accurate estimation of LTI parameters is critical for downstream ablation planning. Agar phantoms were prepared according to Section~\ref{sec:sys_overview} and cured for four hours. An OCT C-scan was then acquired to measure surface inclination with respect to the robot arm. The laser scalpel was aligned with the tissue surface normal, and point ablations were performed at a minimum of three locations for each duty cycle (Fig.~\ref{fig:lti_model}a). The ablation duration was fixed at $1.5$~s across all trials. To ensure high-quality data, the protective window was cleaned between ablations to remove accumulated debris. Debris generation is a well-documented phenomenon in laser tissue ablation, arising from microbubble explosions~\cite{verdaasdonk1990explosive}; debris on the protective window attenuates and diffuses laser energy, and can result in inconsistent craters.  

Following Section~\ref{sec:LTI_model}, LTI parameters were estimated across multiple duty cycles and presented in Table \ref{tb:fitresults}. The super-Gaussian (SG) sharpness parameter $P$ was first treated as a free variable and then fixed to its mean value for subsequent fitting. Given the multimodal profile of the 1060~nm surgical laser (with a 650~nm red alignment beam), the super-Gaussian model consistently outperformed Gaussian fits in capturing crater morphology except in extremely low power cuts, where the Gaussian tip of the crater dominates over the cylindrical body due to the shallow depth resulting from lower power (Fig.~\ref{fig:lti_model}). Final parameters used for simulation are highlighted in blue in Table~\ref{tb:fitresults}. Independent phantom validation yielded an RMSE of $0.402$~mm between measured and simulated data, confirming predictive accuracy. The ablation threshold parameter $\phi$ was determined experimentally as the lowest energy producing a cut (20\% duty cycle, $\phi = 1.939$~J).  

%To validate the LTI model, data from another phantom was obtained and the RMSE between the collected data and simulation using estimated parameters was found to be $0.402\mathrm{mm}$. Note that the parameter $\phi$ was determined by estimating the lowest energy that produced a cut (duty cycle $20\%$, $\phi = 1.939J$).

The SG model consistently achieved lower RMSE values than Gaussian fits, with the greatest improvements observed at higher duty cycles where the multimodal laser profile produced flat-bottom craters and steep sidewalls. On average, the SG model yielded a sharpness parameter of $P = 12.73$, amplitude $A = 0.334 \pm 0.048$, and beam spread $\sigma = 0.473 \pm 0.054$~mm, consistent with quasi-top-hat beam characteristics. These findings demonstrate that the SG formulation provides a more accurate representation of multimodal laser--tissue interactions, enabling improved prediction and planning for volumetric resections.

% Fig.~\ref{fig:lti_model} further illustrates this comparison. 
In Fig.~\ref{fig:lti_model}a, crater morphology is reconstructed from OCT point clouds, with amplitude $A$ and beam spread $\sigma$ describing ablation depth and lateral spread. Fig.~\ref{fig:lti_model}b compares Gaussian and SG fits along a crater cross-section: the Gaussian model underestimates the flat-bottom region, while the SG model closely matches the ground truth, yielding a lower RMSE (0.149 vs.\ 0.326). Fig.~\ref{fig:lti_model}c shows residual distributions, where the SG fit produces a narrower profile, even with average model parameters, confirming improved predictive accuracy.

%\begin{table}[b]
%\centering
%\caption{Parameter Fitting Results and Error for Gaussian and Super Gaussian (SG) Fits}
%\label{tb:errorrmse}
%\begin{tabular}{|c|c||l|l|l|l||l|}
%\hline
%& & SG & \multicolumn{3}{c||}{SG ($P = 12.727$)} & {Gaussian} \\
%\hline
%PWM \% & Trial \# & $P$ & $A$ & $\sigma$ & RMSE & RMSE \\
%\hline\hline
%\multirow{3}*{30} & 1 & 3.38 & 0.428 & 0.390 & 0.141 & \textbf{0.105}\\
% & 2 & 2.67 & 0.344 & 0.431 & 0.119 & \textbf{0.078}\\
%& 3 & 2.98 & 0.341 & 0.379 & \textbf{0.104} & 0.122\\\hline
%\multirow{3}*{50} & 1 & 22.11 & 0.357 & 0.526 & \textbf{0.242} & 0.386\\
% & 2 & 20.74 & 0.333 & 0.519 & \textbf{0.169} & 0.357\\
% & 3 & 10.58 & 0.390 & 0.489 & \textbf{0.199} & 0.326\\\hline
%\multirow{3}*{70} & 1 & 8.97 & 0.270 & 0.524 & \textbf{0.242} & 0.405\\
 %& 2 & 16.63 & 0.252 & 0.517 & \textbf{0.169} & 0.375\\
 %& 3 & 24.06 & 0.295 & 0.487 & \textbf{0.199} & 0.465\\\hline
%\multirow{3}*{90} & 1 & 17.26 & 0.322 & 0.539 & \textbf{0.281} & 0.680\\
 %& 2 & 10.84 & 0.324 & 0.505 & \textbf{0.373} & 0.666\\
% & 3 & 12.52 & 0.350 & 0.492 & \textbf{0.534} & %0.812\\\hline\hline
%\multicolumn{2}{|c||}{Average} & \textcolor{red}{12.73} & \textcolor{red}{0.334} & \textcolor{red}{0.473} & \cellcolor{black} & \cellcolor{black}\\\hline
%\multicolumn{2}{|c||}{Standard Dev.} & 7.52 & 0.048 & 0.054 & \cellcolor{black} & \cellcolor{black}\\\hline
%\end{tabular}
%\end{table}
%%%######################################################
\begin{table}[t]
\centering
\caption{Fitting Results for Gaussian vs. Super-Gaussian (SG) Models}
\label{tb:fitresults}
\resizebox{\columnwidth}{!}{%
\begin{tabular}{|c|c|c|c|c|c|c|}
\hline
\textbf{PWM} & \textbf{Trial} & \textbf{$P$ (SG)} & \textbf{$A$} & \textbf{$\sigma$} & \textbf{RMSE (SG)} & \textbf{RMSE (G)} \\
\hline \hline
30 & 1 & 3.38 & 0.428 & 0.390 & 0.141 & \textbf{0.105} \\
   & 2 & 2.67 & 0.344 & 0.431 & 0.119 & \textbf{0.078} \\
   & 3 & 2.98 & 0.341 & 0.379 & \textbf{0.104} & 0.122 \\
\hline
50 & 1 & 22.11 & 0.357 & 0.526 & \textbf{0.242} & 0.386 \\
   & 2 & 20.74 & 0.333 & 0.519 & \textbf{0.169} & 0.357 \\
   & 3 & 10.58 & 0.390 & 0.489 & \textbf{0.199} & 0.326 \\
\hline
70 & 1 & 8.97  & 0.270 & 0.524 & \textbf{0.242} & 0.405 \\
   & 2 & 16.63 & 0.252 & 0.517 & \textbf{0.169} & 0.375 \\
   & 3 & 24.06 & 0.295 & 0.487 & \textbf{0.199} & 0.465 \\
\hline
90 & 1 & 17.26 & 0.322 & 0.539 & \textbf{0.281} & 0.680 \\
   & 2 & 10.84 & 0.324 & 0.505 & \textbf{0.373} & 0.666 \\
   & 3 & 12.52 & 0.350 & 0.492 & \textbf{0.534} & 0.812 \\
\hline \hline
\multicolumn{2}{|c|}{\textbf{Average}} & \textcolor{blue}{\textbf{12.73}} & \textcolor{blue}{\textbf{0.334}} & \textcolor{blue}{\textbf{0.473}} & \textcolor{blue}{\textbf{0.231}} & \textcolor{blue}{\textbf{0.398}} \\
\multicolumn{2}{|c|}{\textbf{Std. Dev.}} & 7.52 & 0.048 & 0.054 & 0.121 & 0.223 \\
\hline
\end{tabular}
}
\vspace{-10pt}
\end{table}

%%%######################################################
%%%%%%%##################################################
% \begin{align}
%     u &= \left(x_{L}, y_{L}, t_{x}, t_{y}, E\right),
%     \label{eq:laser_shot}\\
%     \Delta z(\mathbf{s}; \mathbf{u}) &= \frac{1}{\rho\,h}\,\max\!\Bigl(0,\;E\,\exp\!\Bigl[-\Bigl(\tfrac{1}{2}\tfrac{r(\mathbf{s})^{2}}{w^{2}}\Bigr)^{P}\Bigr]- \phi \Bigr),
%     \label{eq:crater_depth}
% \end{align}

% where ($x_l$, $y_l$) is the laser beam center at plane $z=0$, and with an angular tilt of $t_x$, and $t_y$ for the given shot energy, $E$ is ???, $s$ is ???, $p$ is ???, $h$ is ???, $\omega$ is ???, and $\phi$ is the ablation enthalpy. 

%%%#########################
\subsection{Volumetric Resection through MPC-Planner}
The surgical planner described in Section~\ref{sec:planner} was evaluated across three increasingly complex tasks: phantom tissue resection with user-defined objectives and constraints, resection with a subsurface critical structure extracted from an OCT C-scan, and lastly a test on \textit{ex-vivo} porcine tissue. All planning tasks were originally simulated in Python, before being translated to the RATS system. Across all experiments, the permissible angular tilt of the laser was limited to $10^{\circ}$ to prevent collisions within the workspace. 

\begin{table}[t]
\centering
\caption{Ablation results comparing feedforward (No Feedback) and closed-loop (Feedback) planning in simulation and real experiments.}
\label{tb:planresults}
\resizebox{\columnwidth}{!}{%
\begin{tabular}{@{}llcccccc@{}}
\toprule
\multicolumn{2}{c}{\textbf{Condition}} & \textbf{Obj. Vol.} ($\mathrm{mm}^3$) & \textbf{RMSE} (mm) & \textbf{MAE} (mm) & \textbf{\%OC} & \textbf{\%UC} & \textbf{IoU} \\
\midrule
\multirow{2}{*}{No Feedback} 
 & Simulation & 71.4 & 0.798 & 0.554 & 15.0 & 24.7 & 62.3 \\
 & Real       & 71.4 & 1.007 & 0.797 & 9.8 & 47.3 & 21.1 \\
\midrule
\multirow{2}{*}{Feedback} 
 & Simulation & 71.4 & 1.199 & 0.842 & 37.1 & 23.3 & 60.0 \\
 & Real       & 71.4 & 0.842 & 0.630 & 19.3 & 25.8 & 60.0 \\
\bottomrule
\end{tabular}
}
\vspace{-10pt}
\end{table}

\begin{figure*}[!htbp]
    \centering
    \includegraphics[width=0.89\textwidth]{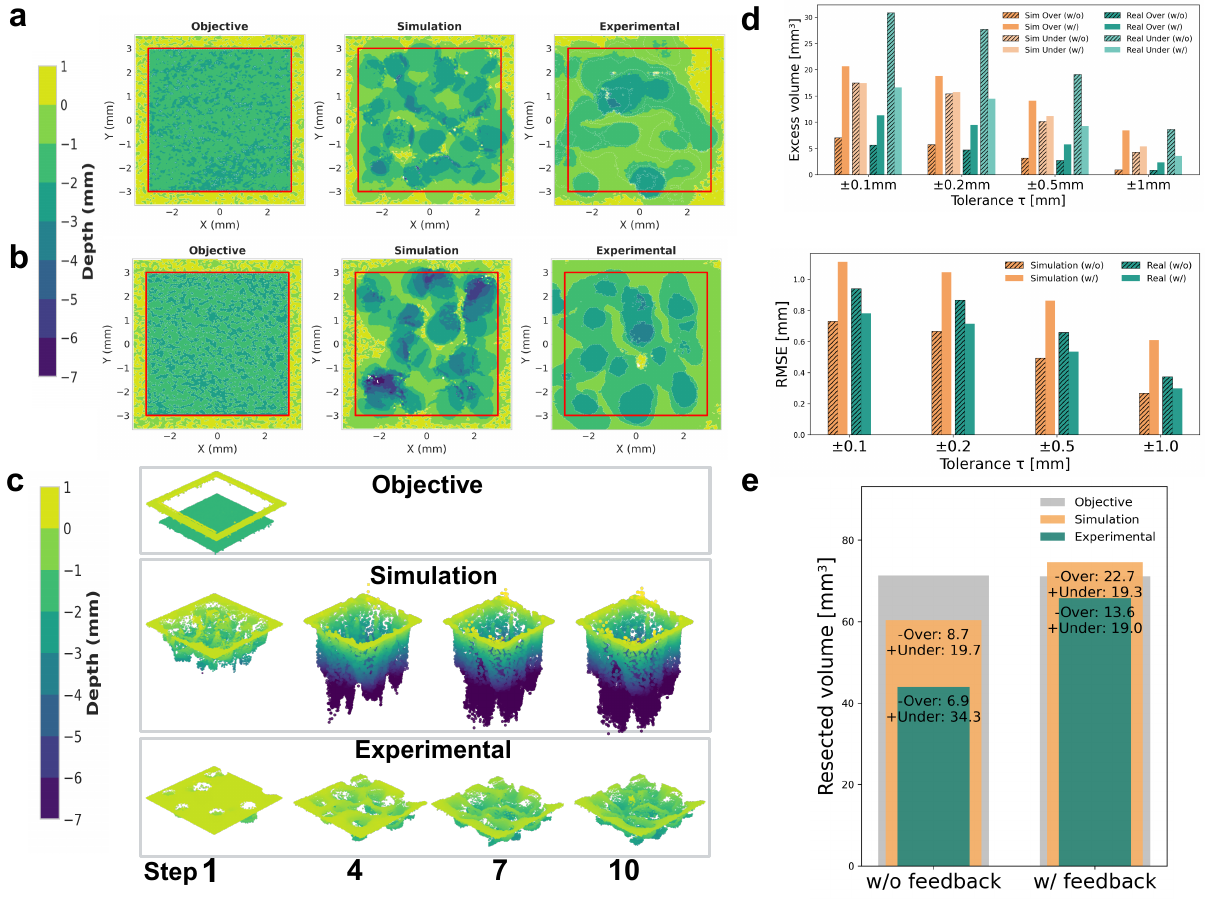}
    \vspace{-0.5em} % adjust spacing
    \caption{\textbf{Volumetric resection planner} \textbf{(a)} without feedback: simulation and experimental result in 2.5D with depth information at each XY point representing color. The red bounding box denotes the objective boundary. The experimental result can be seen undercutting; \textbf{(b)} with repeated OCT feedback. Simulation shows considerable overcut, whereas the experimental result is closer to the objective;
    \textbf{(c)} intermediate steps of (b) showcasing the process of incorporating regular feedback. Experimental resection progresses in controlled manner;
    \textbf{(d)} planner performance in terms of excess volume (top) and RMSE (bottom), over clinical directional tolerance;
    \textbf{(e)} resected volume in comparison to the objective in the simulation and experimental case.
    }
    \label{fig:planning_results}
    \vspace{-5pt}
\end{figure*}

\begin{figure}[!h]
    \centering
    \includegraphics[width=0.9\linewidth, keepaspectratio]{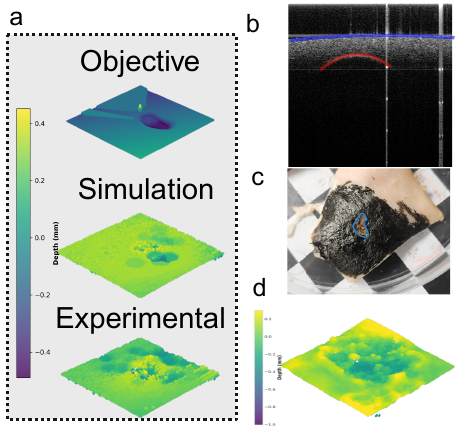}
    \caption{\textbf{Subsurface Resection and Tissue:} \textbf{(a)} Resection objective, simulation, and experimental results for phantom with subsurface critical structure. The objective function is defined by accounting for the constraint. \textbf(b) OCT B-scan showing surface (blue) and constraint (red); \textbf{c} porcine tissue sample post-resection, stained with India Ink dye; \textbf{d} resected result from porcine tissue. }
    \label{fig:planning_demo}
    \vspace{-8pt}
\end{figure}

\subsubsection{Phantom Resection: Feedforward mode}
A $6 \times 6 \times 2~\mathrm{mm}^3$ square well target was chosen and simulated to approximate the typical size of brain tumor samples, where other laser-based thermal therapies, (e.g. laser interstitial thermal therapy) are not applicable. The resection plan was terminated once the residual tumor volume fell below 25\% of the objective. The same plan was then executed on a tissue phantom using RATS, and the window was inspected for debris every three steps (chosen as a heuristic). Upon completion, a post-ablation OCT scan was acquired. The simulated plan, post-ablation volume, and associated errors are presented in Fig.~\ref{fig:planning_results} and Table~\ref{tb:planresults}. The real RMSE and MAE errors increased by $26.19 \%$ and $43.86 \% $ over simulated results. 

\subsubsection{Phantom Resection: Feedback mode} \label{sec:phantom_feedback}
Using the same square well target, an ablation plan was generated and subsequently re-planned every 9 steps to compensate for thermal and debris effects, where 9 steps represents a heuristically chosen balance between execution speed and quality of resection. The final performance metrics are presented in Table~\ref{tb:planresults}. Compared to feedforward, feedback improved IoU agreement between simulation and experiments by $64.8 \%$. 

In the real no-feedback case, a significant undercut was observed, likely caused by attenuation of the laser beam due to loss of focus as the cuts deepened. This effect was mitigated when feedback was incorporated, demonstrating the effectiveness of closed-loop control. The RMSE and MAE error of $1.007$~mm and $0.797$~mm without feedback and $0.842$~mm and $0.630$~mm with feedback confirm that both methods achieve clinically acceptable accuracy within the $1$~mm neurosurgical threshold. Unlike the previous case, when feedback was included, both RMSE and MAE in the real experiment improved (decreased) by $29.77\%$ and $25.17\%$ over the simulation. It should be noted that RMSE values may be inflated by OCT speckle noise. 

%Note that RMSE values could be inflated by OCT speckle noise. 

\subsubsection{Resection with a Subsurface Critical Structure}
Neurosurgical tumors often envelop critical structures such as blood vessels, and thus unnecessary resection can have potentially harmful consequences. Thus, identification and preservation of critical anatomy is a must for a successful resection plan. 
To mimic a subsurface structure, a ball bearing (3.175~mm OD) was embedded beneath the surface of a phantom, chosen for its optical distinctiveness. The bearing was segmented from the OCT scans using the DBSCAN algorithm \cite{ester1996density} for clustering. The segmented subsurface structure was isolated and designated as a hard constraint to avoid during planning. The phantom surface was uneven, resembling tissue-like variability (Fig.~\ref{fig:planning_demo}a, b). Unlike the previous square-well experiment, an artificial tumor shape was defined using an irregular closed spline in the $XY$-plane centered at the tumor location, with mean radius $1.5$~mm and random radial jitter to produce a non-circular footprint. The maximum resection depth was limited to $2 \mathrm{mm}$. A constraint boundary was developed by interpolation using the Clough-Toucher algorithm, and a clearance margin of $0.2$~mm was added to the constraint to enforce safety. The objective function was generated following the constraints and a resection plan was generated with feedback and executed. As seen in Fig.~\ref{fig:planning_demo}a, the embedded subsurface structure was within $0.5$mm of the top surface, resulting in the planner making smaller cuts. 

\subsubsection{\textit{Ex vivo} tissue resection}
Finally, the system was evaluated on an \textit{ex-vivo} porcine abdominal tissue to qualitatively demonstrate the system capability for real-tissue resection (Fig.~\ref{fig:planning_demo}c, d). %A final demonstration was done to simulate surgical complexity by extending the planning method to \textit{ex vivo} porcine tissue extracted from the abdomen. 
The surface was coated with an absorbing chromophore (India Ink) to lower the ablation threshold to 1060 nm and induce initial ablation. The addition of India Ink is used for \textit{ex-vivo} and \textit{in-vitro} experiments due to the low natural absorption of tissue specimens at 1060 nm. For clinical implementation, the usage of alternative resection methods (e.g. 10600 nm CO2 lasers) would bypass the requirement for tissue absorption augmentation. Using phantom-derived LTI parameters without retraining, RATS achieved a $4 \times 4 \times 2~\mathrm{mm}^3$ resection.
%
% Ablation of tissue generates smoke alongside microbubble explosion, further covering the protective window.
%
%A target of size $4\times4\times2$ mm$^3$ is used for demonstration purposes. %The pre-and-post resection surface is shown in Figure \ref{fig:planning_demo}. 
While accuracy was reduced compared to phantom trials due to debris, scattering, and a lack of a porcine-specific LTI model, the experiments demonstrated feasibility and offered critical insights into the difficulties of working with \textit{ex-vivo} tissue. These findings highlight robustness under anatomically realistic conditions and motivate the development of tissue-specific LTI models. Note that the above performance can be improved at the cost of additional runtime and a tissue-specific LTI model.

\section{Discussion And Conclusion}
%%%######################################################
This work introduces RATS, the first closed-loop robotic platform for OCT-guided volumetric laser resection. By integrating OCT and RGB-D imaging, sub-millimeter multi-stage calibration, and a sampling-based MPC planner with closed-loop feedback, RATS achieved clinically relevant performance with a calibration accuracy of $0.161 \pm 0.031$~mm, LTI modeling (RMSE $0.231 \pm 0.121$~mm), and volumetric resection accuracy of $0.842$~mm RMSE, improving IoU by 64.8\% compared to feedforward execution, while preserving subsurface structures in phantom and \textit{ex-vivo} studies. 

% Our results highlight the possibility of obtaining clinically relevant performance with practical hardware design, an empirical LTI model, and a modular sampling-based MPC planner for volumetric resection. The proposed calibration methods extend beyond the proposed application to other free-form or fiber-coupled laser systems. The tissue phantom developed is from easy-to-source, biocompatible materials to enhance the laser scalpel's performance while accounting for suitability with OCT-based scanning. The entire codebase is implemented in Python on an Intel i9 CPU (11th gen, Intel Corporation, CA, USA) and has the potential to be further sped up with a C-based implementation in the future. 

Our results highlight novel yet practical hardware design, an empirical LTI model, and a modular sampling-based MPC planner for volumetric resection. The method is device and laser-agnostic and can be adapted to any system with a tissue point cloud, LTI model, and controllable laser power and orientation. The proposed calibration methods generalize to other free-form or fiber-coupled laser systems and our tissue phantom is developed from easy-to-source, biocompatible materials while remaining compatible with both laser and OCT. The codebase is implemented in Python on an Intel i9 CPU (11th gen, Intel Corporation, CA, USA) and can be further sped up with a C-based implementation in the future. 

Despite these advances, several limitations remain. Debris accumulation during ablation reduces laser transmission and accuracy, motivating the integration of air flushing and extended working distances. The current laser spot size ($0.9$ mm) limits precision; improved optics targeting $<0.5$ mm are required for fine resections. Lastly, the LTI model is geometric and does not account for thermal accumulation across repeat ablations, which can alter tissue properties (Fig.~\ref{fig:planning_results}e). Finally, the sampling-based planner is slower than raster strategies, though it enables constraint-aware resections in geometrically complex cases.

In conclusion, we present RATS, a novel robotic platform integrating OCT imaging and laser-based resection for intelligent, precise, and safe surgical intervention. 
By addressing hardware, modeling, and control challenges, RATS demonstrates the potential of autonomous, constraint-aware laser resections in neurosurgery and oncology, advancing surgical robotics toward \textit{in-vivo} tumor applications. This work advances the frontier of autonomous surgical robotics and lays the foundation for translation toward \textit{in-vivo} tumor resections. Future work will undertake improvements in opto-mechanical design for surgical deployment, a generalizable LTI model \cite{ma20233d}, and near real-time adaptive planning for achieving sub-mm surgical performance.

\section*{Acknowledgments}
The authors wish to thank Aislinn Hurley and the members of the Brain Tool Lab for support and feedback.

\bibliographystyle{IEEEtran}  % e.g., plain, apalike, IEEEtran
\bibliography{IEEEfull}  % no .bib extension

\end{document}